\PassOptionsToPackage{table}{xcolor}
\documentclass[10pt,twocolumn,letterpaper]{article}

\usepackage[pagenumbers]{cvpr} % To force page numbers, e.g. for an arXiv version

\usepackage[dvipsnames]{xcolor}

\definecolor{cvprblue}{rgb}{0.21,0.49,0.74}
\usepackage[pagebackref,breaklinks,colorlinks,citecolor=cvprblue]{hyperref}
\usepackage{multirow}
\usepackage{pifont}
\usepackage{enumitem}
\usepackage{xcolor}

\definecolor{mygray}{gray}{.9}

\def\paperID{2560} % *** Enter the Paper ID here
\def\confName{CVPR}
\def\confYear{2024}

\title{Edit3K: Universal Representation Learning for Video Editing Components}
\author{Xin Gu$^{1,2,\dagger, \ddagger}$, Libo Zhang$^{2,3,\dagger}$, Fan Chen$^{1}$, Longyin Wen$^{1}$, Yufei Wang$^{1}$, Tiejian Luo$^{2}$, Sijie Zhu$^{1,*}$ \\
$^{1} $ByteDance Inc. $^{2}$University of Chinese Academy of Sciences \\ $^{3}$Institute of Software Chinese Academy of Sciences\\
{\tt\small \{fan.chen, longyin.wen, yufei.wang, sijiezhu\}@bytedance.com} \\ \tt\small guxin21@mails.ucas.edu.cn, libo@iscas.ac.cn,  tjluo@ucas.ac.cn
}

\begin{document}
\maketitle
\let\thefootnote\relax\footnotetext{$*$ Corresponding author, sijiezhu@bytedance.com}
\let\thefootnote\relax\footnotetext{$\dagger$ Equal contribution}
\let\thefootnote\relax\footnotetext{$\ddagger$ This work was done during the first author’s internship at ByteDance}

% \input{sec/abstract}  
% \input{sec/introduction} 
% \input{sec/related_work} 
% % % \input{sec/task} \Sijie{Maybe no need this part}
% \input{sec/dataset} 
% \input{sec/method} 
% \input{sec/experiment} 
% \input{sec/conclusion} 

% =========================================================
\begin{abstract}
 This paper focuses on understanding the predominant video creation pipeline, \ie, compositional video editing with six main types of editing components, including video effects, animation, transition, filter, sticker, and text. In contrast to existing visual representation learning of visual materials (\ie, images/videos), we aim to learn visual representations of editing actions/components that are generally applied on raw materials. We start by proposing the first large-scale dataset for editing components of video creation, which covers about $3,094$ editing components with $618,800$ videos. Each video in our dataset is rendered by various image/video materials with a single editing component, which supports atomic visual understanding of different editing components. It can also benefit several downstream tasks, \eg, editing component recommendation, editing component recognition/retrieval, etc. Existing visual representation methods perform poorly because it is difficult to disentangle the visual appearance of editing components from raw materials. To that end, we benchmark popular alternative solutions and propose a novel method that learns to attend to the appearance of editing components regardless of raw materials. Our method achieves favorable results on editing component retrieval/recognition compared to the alternative solutions. A user study is also conducted to show that our representations cluster visually similar editing components better than other alternatives. Furthermore, our learned representations used to transition recommendation tasks achieve state-of-the-art results on the AutoTransition dataset. The code and dataset are released at \url{https://github.com/GX77/Edit3K}.
% \keywords{Video Editing Component \and Representation Learning}
\end{abstract}

\section{Introduction}
\label{sec:introduction}

\begin{figure}
    \centering
    \includegraphics[trim=0 0 0 0, clip, width=\linewidth]{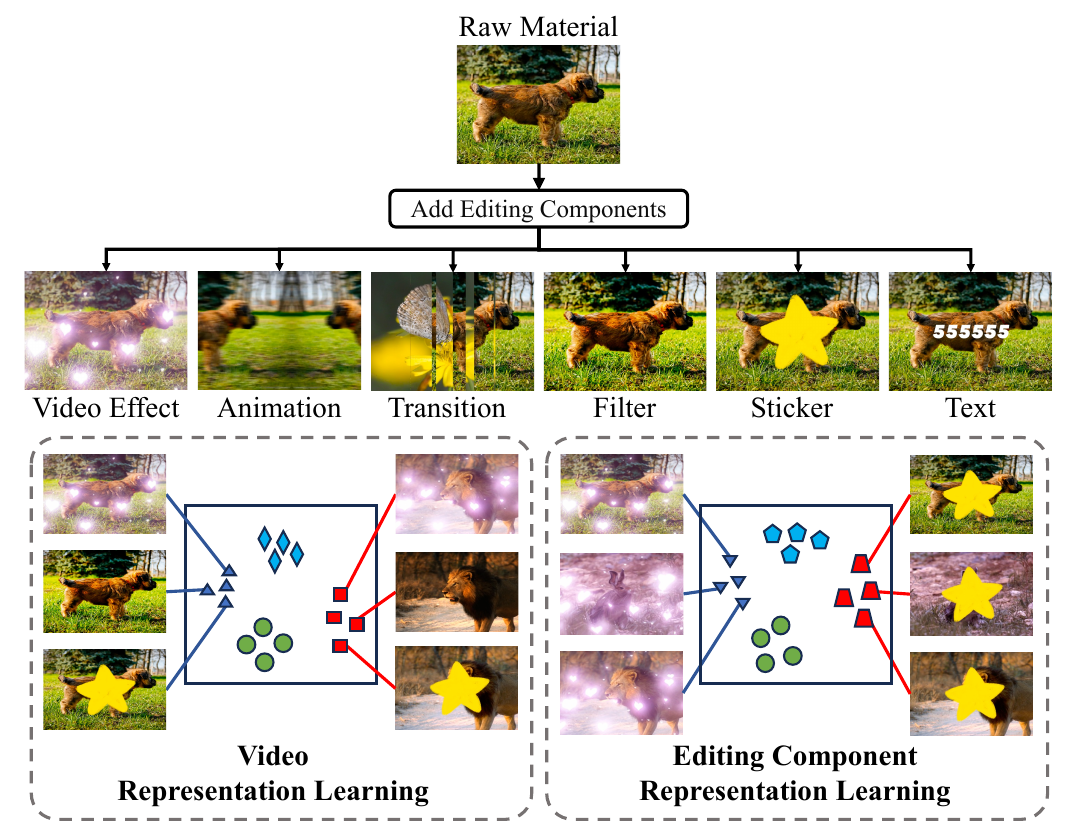}
    %\vspace{-0.6cm}
    \caption{An overview of generic video representation learning and editing components representation learning. The embedding of generic video representation learning is clustered based on video content, \eg, semantics, context, etc, while the ideal editing component representation should be only dependent on the applied editing components rather than the content of raw materials.} % Best viewed on screen with zoom-in.
\label{fig:intro}
\end{figure}

% Video creation has become the major form of data for social media, advertising, education, entertainment.
Video has emerged as a major modality of data across various applications, including social media, advertising, education, and entertainment. The predominant video creation pipeline for production, \ie, compositional video editing, is based on the combination of editing components (Fig. \ref{fig:intro}), \eg, video effect, animation, transition, filter, sticker, and text. Tons of videos are created with these editing components every day on various video creation platforms, but little effort has been made to understand these editing components. Specifically, learning universal representations for editing components is unexplored but critical for lots of downstream tasks in video creation fields, \eg, effects recommendation, detection, recognition, generation, etc. 
%Fig. \ref{fig:intro} shows a typical video creation pipeline and an example of the 6 types of editing components.

%A typical video editing pipeline first creates tracks and slots as temporal arrangements of different materials. The materials are then assigned to specific slots, and the materials could be pre-processed with editing actions, \eg spatial/temporal cropping. For each slot, editing actions/components will be applied to the material so that the rendered result is more visually pleasing, \eg adding heart-shaped video effects to make it romantic. 
% Video creation pipeline and definition of editing components. //

% \begin{table}[!htbp]
%     \centering
%     \begin{tabular}{c c c c}
%     \hline

%     \hline
%          & Motion & Local Appearance & Global Appearance & Additional Layer \\
%     \hline
%     Video Effect  & Additional layer with new content, Motion of the material, \\
%     Animation &   \\ 
%     Transition &   \\
%     Filter &  \\
%     Sticker & \\
%     Text (Font) &  \\
%     \hline

%     \hline
%     \end{tabular}
%     \caption{Definition of the 6 major types of editing components.}
%     \label{tab:definition}
% \end{table}
Existing video representation learning methods \cite{feichtenhofer2019slowfast, tong2022videomae, he2022masked, wei2022masked, wang2023videomae} are usually developed on object-centric action recognition datasets, \eg, Kinetics \cite{kay2017kinetics}, to encode the information from the video content, \eg, semantics, the action of the subjective, context, motion, etc. However, editing components usually do not have clear semantics, subjective action, or context information. They could be a simple chroma change of the whole image or local effects like scattering stars. Some components are special homography transformations on the raw material (\ie, image/video) and the rendered result is highly dependent on the appearance of the raw material. Thus, it is extremely challenging to learn a representation that encodes the information of editing components regardless of raw materials. Fig. \ref{fig:intro} shows a comparison between generic video representation learning and our task. To the best of our knowledge, how to learn universal representations for diverse types of editing components remains unexplored in representation learning literature, and none of the existing datasets supports research on learning universal representation for the 6 major types of video editing components.
% In this address 
% Challenges for this problem. 1) address with a new dataset. 2) limitation of existing visual representation learning. Must address with a novel method.

% In this paper, we address these challenges and limitations. Our design.... Our evaluation. State-of-the-art performance.

In this paper, we propose the first large-scale video editing components dataset, \ie, Edit3K, to facilitate research on editing components and automatic video editing. The proposed dataset contains $618,800$ videos covering $3,094$ editing components of 6 major types. Each video is rendered with one editing component with both image and video materials to enable atomic research on single editing components. Given that the editing component is visually mixed with raw materials, the key challenge of this problem is to distinguish editing components in the rendered frames but ignore the changes in the raw materials. We benchmark popular representation learning methods, \eg, contrastive learning and masked autoencoder, and propose a tailored contrastive learning loss that better solves this problem. In addition, we propose a novel embedding guidance architecture that enhances the distinguishing ability of editing components by providing feedback from the output embedding to the model. The learned editing element representations of our method can be directly applied to downstream tasks \eg, transition recommendation \cite{shen2022autotransition}, achieving state-of-the-art results. Attention map visualization shows that our model learns to focus on the editing components without pixel-level supervision. We summarize our contributions as follows:
\setlist{nolistsep}
\begin{itemize}[noitemsep,leftmargin=*]
\item We introduce the first large-scale dataset for video editing components covering $3,094$ atomic editing actions of 6 major categories, which makes it possible to learn universal representations for all 6 categories of editing components.
% It can benefit the understanding of video editing and downstream tasks, \eg recommendation, etc. 
\item We propose a novel embedding guidance architecture that aims to distinguish the editing components and raw materials, along with a specifically designed contrastive learning loss to guide the training process.  
\item We benchmark major alternative methods and validate the superiority of our method in various scenarios. Extensive experiments show state-of-art performance on both Edit3K and AutoTranstion datasets. 
\end{itemize}

% Furthermore, we propose a contrastive-learning-based method with different supervision from the existing representation learning methods. We assign videos with the editing components as positive samples, and videos with different editing components as negative samples, so that the model is sensitive to editing components but not sensitive to the material changes. To enhance the distinguishing ability of the 

%===========================================

\section{Related Work}
\label{sec:related}

\textbf{Video Editing/Creation}
Conventional workflow of video editing requires a deep understanding of the raw image/video materials and professional knowledge of editing and aesthetics, \eg, manually cutting videos and selecting editing effects, thus is extremely time-consuming for the video creators. Automatic video editing can significantly improve editing efficiency and it has been explored from different perspectives \cite{frey2021automatic, koorathota2021editing, shen2022autotransition}. Koorathota \etal \cite{koorathota2021editing} focus on contextual and multimodal understanding for video editing. Frey \etal \cite{frey2021automatic} transfer editing styles from a source video to a target video with matched footage, considering framing, content type, playback speed,
and lighting of input video segments. AutoTransition \cite{shen2022autotransition} proposes to automatically recommend transitions based on the previous video frames and audio. Recently, there has been a thread of works \cite{ceylan2023pix2video, yang2023rerender, liew2023magicedit} using GAN \cite{goodfellow2020generative} or diffusion \cite{rombach2022high,ho2020denoising} model to generate videos, but they are still not widely deployed in real-world video creation applications. \textit{We consider them as orthogonal works and they could be combined with our work in the future.}\\
\noindent\textbf{Editing Components for Video Creation}
To the best of our knowledge, none of the existing works provides a comprehensive study on video editing components. The most related work, AutoTransition \cite{shen2022autotransition} is proposed for a specific downstream task, \ie, transition recommendation. The AutoTransition dataset contains 104 transitions and only 30 transitions are used for the transition recommendation task. \textit{This work only focuses on one specific type and does not support learning universal representation for all the major 6 types of editing components.}\\ %, and our dataset is the first one to address this limitation. 
%  Sticker820K \cite{zhao2023sticker820k} contains about 820k static stickers, \ie each sticker is an image, and the corresponding text annotations.
\noindent\textbf{Visual Representation Learning} Visual representation learning has been extensively studied for video data \cite{feichtenhofer2021large}. Early works conduct supervised classification \cite{tran2019video, feichtenhofer2019slowfast} on large-scale datasets, \eg, Kinetics \cite{kay2017kinetics}, to learn strong pre-trained representation. Recent self-supervised representation learning methods, \eg, contrastive learning \cite{oord2018representation, he2020momentum, pan2021videomoco, chen2021exploring} and masked autoencoder \cite{he2022masked, tong2022videomae,arnab2021vivit}, have achieved comparable performance as supervised methods on downstream tasks\cite{feichtenhofer2021large}. The key insight is to learn a representation space where visually similar videos are close to each other. \textit{However, existing works have limitations on learning representation for editing components from video data (Sec. \ref{sec:exp_main}), because these methods are not designed to distinguish the raw material and editing components in the video frames.}

\begin{figure*}[!htbp]
 \centering
 \includegraphics[width=\linewidth]{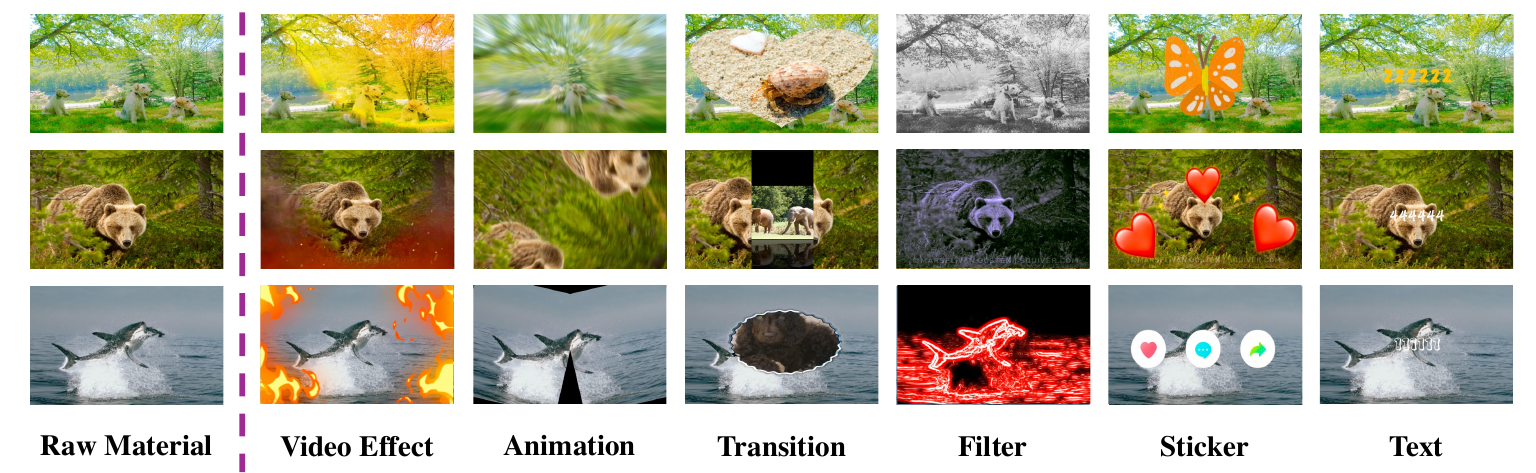}
 \caption{Examples of 6 major types of video editing components, \ie, video effect, animation, transition, filter, sticker, and text.}
 % \vspace{-0.15cm}
 \label{fig:example}
\end{figure*}

\begin{table*}[!htbp]
\centering
% \renewcommand{\arraystretch}{1.2}
% \resizebox{\linewidth}{!}{
\begin{tabular}{l |cccccc | c}
\rowcolor{mygray}
\specialrule{1.5pt}{0pt}{0pt}
& Video Effect & Animation & Transition &  Filter & Sticker & Text & \textbf{Total} \\ 
\hline
Classes & 888 & 176 &  204 &  228 & 1000 & 598 & \textbf{3,094} \\
Rendered Videos & 177,600  & 35,200 & 40,800 & 45,600 &200,000 & 119,600 & \textbf{618,800} \\ 
\specialrule{1.5pt}{0pt}{0pt}
\end{tabular}
% }
\caption{The statistical information of the proposed Edit3K dataset.}
% \vspace{-0.15cm}
\label{tab:dataset1}
\end{table*}

\section{Edit3K Dataset}
\label{sec:dataset}
Existing video datasets mostly focus on understanding the overall content of the video frames, \eg, action, context, semantics, and geometry, which can not support the research for understanding video editing components. The most related dataset is proposed in AutoTransition \cite{shen2022autotransition}, but this dataset only contains one editing component type (\ie, 104 transitions) out of the major six types with limited coverage. To the best of our knowledge, our dataset is the first large-scale editing component dataset for video creation covering all the six major types, \ie, video effect, animation, transition, filter, sticker, and text. 

% \begin{figure}
%     \centering
%     % \includegraphics{}
%     \caption{Examples of 6 major types of video editing components, \ie video effect, animation, transition, filter, sticker, and text.}
%     \label{fig:example}
% \end{figure}

\subsection{Editing Component Definition}
There is no formal definition for the 6 types of editing components in the academic literature, so we summarize the major difference between them and describe their connection with existing concepts. Examples are shown in Fig. \ref{fig:example}.

Both video effects and filters change the appearance of material images or video frames toward a specific style, but video effects focus more on local editing, \eg, adding shining star-shaped markers. Filters mainly change the overall style of the whole scene, \eg, adjusting the illumination and chroma. Animation can be considered as a homography transformation applied to the whole material, which is similar to camera viewpoint changes. The transition is applied to connect two materials with appearance changes and fast motion. The sticker simply uses copy-paste to add another foreground object on top of the current material, which is similar to image composition \cite{zhang2021deep}. However, some of the stickers are video stickers and the stickers may look different in each frame. Text is a special sticker whose content can be edited by the user, but the text style can be changed by applying different fonts. We focus on text font/style rather than the content in this paper.

\subsection{Dataset Generation}
It is challenging to collect an academic dataset for editing components because online published videos \cite{shen2022autotransition} usually apply multiple editing components sparsely in an entangled way. Most of the frames may contain no editing component or only a simple filter, while a small number of frames (\eg a $0.5$ s slot) could have stickers/text, video effects, and animations at the same time. To address this limitation, we propose to render videos based on existing raw materials and a pre-defined set of editing components using a free-to-use video editing tool, \ie, \textit{CapCut} \cite{capcut}. Each video only contains one disentangled editing component to support atomic research on all editing components with a balanced distribution. 

We use images and videos from the ImageNet \cite{deng2009imagenet} and ImageNet-Vid \cite{deng2009imagenet,shang2017video} datasets as raw materials for rendering. For each video, we randomly take two images/videos as the source material pair for two video slots, and each slot lasts 2 seconds. If the component is a transition, we add it between the two slots, which lasts for 2s, and each slot is reduced to 1s. Otherwise, we apply the component on both the two slots, which covers 4s. In total, we generate $618,800$ videos covering $3,094$ editing components.

\begin{figure*}[!htbp]
 \centering
 \includegraphics[width=\linewidth]{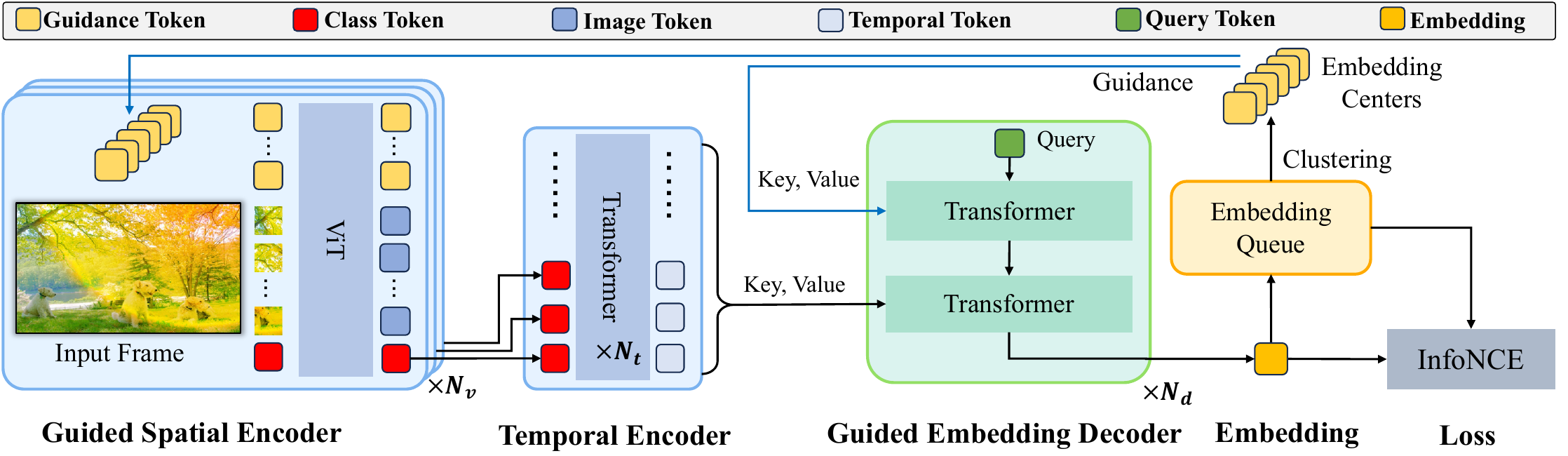}
 \caption{An overview of the proposed method. The input video frames are fed to the spatial and temporal encoder to generate the visual features. Then the embedding decoder takes the visual features as key, value and generates the final editing component embeddings using cross-attention mechanism with one query token. All encoders and decoders are guided with guidance tokens which are the embedding centers of the embedding saved in a queue. The model is optimized with InfoNCE \cite{oord2018representation, he2020momentum} loss across the batch and the embedding queue provides extra negative samples for an extra loss term. Best viewed on screen with zoom-in.}
 \label{fig:framework}
\end{figure*}

\subsection{Dataset Statistics}
% \Sijie{Compare with other editing element dataset, e.g. Yaojie's. Compare with video action recognition datasets.}
Tab.~\ref{tab:dataset1} shows the detailed statistical information of the 6 categories of editing components in the proposed dataset.
% We compare our dataset with major existing video datasets in Tab.~\ref{tab:dataset2}.
Our dataset contains significantly ($29.75\times$) more editing components than AutoTransition \cite{shen2022autotransition} dataset. Unlike AutoTransition \cite{shen2022autotransition} which adopts online published videos with unbalanced distribution, our dataset is more balanced across different editing components which is critical for representation learning. Our dataset is the first one to provide editing components with static image materials, which provides valuable research source data for understanding the motion effect of editing components.

% Our dataset is also comparable to major action recognition datasets in terms of  number of videos and total duration. It can be used for self-supervised pre-training or generalization evaluation of foundation models.

%\begin{figure}
%    \centering
%    \includegraphics[trim=0 0 0 0, clip, width=0.5\linewidth]{Pie.pdf}
    %\vspace{-0.6cm}
%    \caption{\redtext{The number of classes for each editing element.}}
%\label{tab:dataset1}
%\end{figure}

% ===================================================

\section{Method}
\label{sec:method}
Fig.~\ref{fig:framework} shows an overview of the proposed method. We first formulate the problem in Sec.~\ref{sec:method-formulation}. Then we describe the guided spatial-temporal encoder in Sec.~\ref{sec:method-encoder} and the guided embedding decoder in Sec.~\ref{sec:method-decoder}. Sec.~\ref{sec:embeddding queues} introduces the embedding queue mechanism and Sec. \ref{sec:method-loss} describes the loss function.

\subsection{Problem Formulation}
\label{sec:method-formulation}
Given the input video frames $\mathcal{F} = \{f_i\}_{i=1}^{N_{v}}$ where $f_i \in \mathbb{R}^{H \times W \times C}$ ($H,W,C$ denote the height, width and channels), the goal of our model is to generate an embedding that encodes the information of the editing component in the video for recognition, recommendation, etc. An ideal embedding should be dependent on the applied editing components but independent of the raw video materials. In addition, we expect visually similar editing components to be close to each other in the embedding space, \eg, all heart-shaped video effects may be distributed in a cluster. 

To achieve this goal, we formulate this special representation learning task as a contrastive learning problem. Given a set of raw material videos $\mathcal{M}=\{m_{i}\}_{i=1}^{N_{m}}$ and editing components $\mathcal{E}=\{e_{j}\}_{j=1}^{N_{e}}$, we can render each raw video with all the $N_{e}$ editing components, resulting in $N_{m}\times N_{e}$ rendered videos $\{(m_{i}, e_{j})\}$. For the editing component $e_k$, all the rendered videos using this editing component $\{(m_{i}, e_k)\}$ are considered as positive samples for each other. On the other hand, all the rendered videos with other editing components $\{(m_{i}, e_{j})|j \neq k\}$ are considered negative samples for $e_k$. A contrastive loss \cite{oord2018representation, he2020momentum} (Sec. \ref{sec:method-loss}) is then applied to pull the positive samples closer while pushing the negative samples away in the embedding space.
% {\noindent \textbf{Visual tokens.}} For each edited video, we first uniformly sample video frames $\mathcal{F} = \{f_i\}_{i=1}^{N_{v}}$, where $f_i \in \mathbb{R}^{H \times W \times C}$. The $H$ and $W$ are the resolution of the video frame and $C$ is the number of channels. Then, each video frame is evenly divided into image patches $\mathcal{I} \in \mathbb{R}^{N_{p} \times (P^2 \times C)}$, where $P$ is the resolution of each image patch, and $N_p = \frac{H \times W}{ P^2}$ is the resulting number of patches. 

% {\noindent \textbf{Guidance tokens.}} We obtain embedding centers of the six major types from the editing component embedding queue (details in Section 4.4), are denoted as $\mathcal{E} = \{e_{vi}, e_{an}, e_{tr}, e_{fi}, e_{st}, e_{te}\}$, where $\mathcal{E} \in \mathbb{R}^{6 \times C}$. 

% {\noindent \textbf{CLS tokens.}} In order to extract the features of each frame, we initialized a special token for each frame, are denoted as $\mathcal{T} = \{t_i\}_{i=1}^{N_v}$, where $\mathcal{T} \in \mathbb{R}^{N_v \times C}$.

\subsection{Guided Spatial-Temporal Encoder}
\label{sec:method-encoder}
\noindent \textbf{Guided Spatial Encoder.} As shown in Fig.~\ref{fig:framework}, the $N_{v}$ inputs frames are fed to the spatial encoder separately. We follow ViT \cite{dosovitskiy2020image} to evenly divide the input frame into small patches with the size of $32 \times 32$. The patches are fed to a linear projection layer \cite{dosovitskiy2020image} to generate patch embedding, and positional embedding \cite{dosovitskiy2020image} is added to each embedding to generate the image tokens (blue squares in Fig. \ref{fig:framework}). 
To help the model distinguish editing components and raw materials in the rendered frames, we add another set of guidance tokens (yellow squares in Fig.~\ref{fig:framework}) in the input, which will be described in Sec. \ref{sec:embeddding queues}. \textit{The guidance tokens can be considered as feedback from the learned embedding to the input and provide the spatial encode with prior knowledge of possible editing components.} The class token (red square in Fig. \ref{fig:framework}) is concatenated to the input tokens to aggregate the information from all the tokens. The tokens are fed to multiple transformer layers with multi-head self-attention \cite{vaswani2017attention}, and the output class token of the last transformer layer is used as the output of the whole frame. 

\noindent \textbf{Temporal Encoder.} For $N_{v}$ input frames, $N_{v}$ output class tokens are generated separately without temporal correlation. However, some editing components, \eg, animation, and transition, contain strong motion information which requires strong temporal information to understand. Therefore, we add a temporal encoder containing $N_{t}$ self-attention \cite{vaswani2017attention} transformer blocks to learn the temporal correlation between frames. In addition, some editing components may be indistinguishable if the sequential information is missing. For example, ``move to left transition" played in reverse order is the same as ``move to right transition", so we add position embedding \cite{vaswani2017attention} to the input tokens of the temporal encoder to provide sequential information.

\subsection{Guided Embedding Decoder}
\label{sec:method-decoder}
The output tokens of the spatial-temporal encoder contain the mixed information from the editing components and the raw materials. Inspired by end-to-end object detection DETR~\cite{carion2020end}, we leverage the cross-attention \cite{vaswani2017attention} mechanism to extract information corresponding to the editing components from the input tokens. As shown in Fig. \ref{fig:framework}, we first adopt the guidance tokens (Sec. \ref{sec:embeddding queues}) as the (key, value) tokens of the cross-attention transformer block \cite{vaswani2017attention}, which represent the historical embedding of editing components. One query token is fed to the first transformer block to extract prior knowledge of the editing components embedding. The output token is then fed to the second transformer block as the query token and the output of the spatial-temporal encoder is used as (key, value) tokens. The guided embedding decoder contains $N_{d}$ layers of these two transformer blocks, and the output token of the last layer is used as the final embedding of the input video.

\subsection{Embedding Queues}
\label{sec:embeddding queues}
The limited batch size is not able to provide enough hard negative samples for contrastive learning and this issue is typically addressed with sample mining \cite{wu2017sampling} or memory bank/queue \cite{he2020momentum}. Different from the memory bank in MoCo \cite{he2020momentum}, we build the dynamic embedding queues to save the recently generated embedding corresponding to all editing components instead of the whole video set. For each specific editing component in the training set, we maintain a first-in-first-out (FIFO) queue with the size of $5$ to save the most recently generated embedding corresponding to this editing component during training. The memory cost of the embedding queues is negligible, but it provides a large number of negative samples for contrastive learning (Eq. \ref{eq:loss_2}). All the embeddings are $l_{2}$-normalized before joining the queue.

Moreover, the embedding queues provide prior knowledge of all the editing components that can be used as guidance to improve the spatial-temporal encoder and the embedding decoder for distinguishing editing components from raw videos. However, using thousands of embedding as guidance tokens would cost a lot in terms of GPU memory and computation, we thus adopt the embedding centers as the guidance tokens. 
Since the 6 types of editing components are naturally clustered into 6 corresponding centers in the embedding space, we compute the embedding centers for the 6 types as guidance tokens by default, which involves negligible memory and computation costs. Alternatively, k-means clustering centers also perform well as guidance tokens (Sec. \ref{sec:ablation}).
% \redtext{How to get the embedding centers? We first apply the k-means algorithm to cluster all the editing elements and then compute the center of the embedding vectors for each cluster.}

% \Sijie{Embedding centers}

\subsection{Loss Function}
\label{sec:method-loss}
% \Sijie{In-batch InfoNCE, embedding queue InfoNCE}
Our model is optimized with two loss terms, \ie, in-batch loss and embedding queue loss. We adopt the widely used InfoNCE loss \cite{oord2018representation, he2020momentum} formulation but use our task-specific positive and negative definition (Sec. \ref{sec:method-formulation}). We first sample $N_{b}$ editing components in a batch, \ie, $\{e_{i}\}_{i=1}^{N_{b}}$. For each editing component $e_{i}$, we randomly select two positive samples, and their embeddings are denoted as $q_{i}$ and $k_{i}$. Apparently, other embeddings in this batch, \eg, $\{k_{j}\}_{j\neq i}$ are negative samples for $q_{i}$ because they correspond to different editing components. The in-batch loss is written as:
\begin{equation}
\mathcal{L}_{batch} = \frac{1}{N_{b}}\sum^{N_{b}}_{i=1} -log\frac{exp(q_{i}\cdot k_{i}/\tau)}{\sum^{N_{b}}_{j=1}exp(q_{i}\cdot k_{j}/\tau)},
\label{eq:loss_1}
\end{equation}
where $\cdot$ means the cosine similarity operation between two embeddings and $\tau$ is the temperature. Assume that we have $N_{e}$ editing components in total for the training. There are $N_{e}$ embedding queues (Sec. \ref{sec:embeddding queues}) saving the most recent embeddings generated during training, we take the $l_{2}$-normalized average embedding in each queue as the reference embedding, \ie, $\{r_{j}\}_{j=1}^{N_{e}}$. The embedding queue loss term is given as:
\begin{equation}
\mathcal{L}_{queue} = \frac{1}{N_{e}}\sum^{N_{e}}_{i=1} -log\frac{exp(q_{i}\cdot r_{i}/\tau)}{\sum^{N_{e}}_{j=1}exp(q_{i}\cdot r_{j}/\tau)}.
\label{eq:loss_2}
\end{equation}
This term covers other hard negative editing components for $q_{i}$ and improves both the performance and training stability. The final loss is computed as $\mathcal{L} = \mathcal{L}_{batch} + \mathcal{L}_{queue}$.
% The InfoNCE loss function is used to guide the training of our method. Specifically, we calculated two losses, which are In-batch InfoNCE and embedding queue InfoNCE respectively.

% {\noindent \textbf{In-batch InfoNCE Loss.}} During the training process, the model randomly samples a batch of videos from the dataset each time and outputs the embedding of the editing component from each video, and then calculates the in-batch infoNCE loss between them. Videos with the same type of editing component are considered positive samples for each other; otherwise, they are considered negative samples.

% \begin{equation*}
% \label{eq:loss}
%     L = -\frac{1}{N} \sum_{i=1}^{N} \log \frac{\exp(s(v_i, v_{+}) / \tau)}{\exp(s(v_i, v_{+}) / \tau) + \sum_{j=1, j\neq +}^{N}\exp(s(v_i, v_{j}) / \tau)}
% \end{equation*}

% where $s(v_i, v_j)$ denotes the similarity between the embeddings of editing component $v_i$ and $v_j$. The temperature parameter $\tau$ is used to adjust the scale of this similarity. $N$ represents the number of samples in the batch. For the sample pairs, $v_i$ and $v_j$ form a positive pair, while $v_i$ and $v_k$ constitute a negative pair.

% {\noindent \textbf{Embedding Queue InfoNCE Loss.}} We also use the editing embeddings in embedding queue to supervise model training, that is, we calculate the infoNCE loss between the embeddings extracted from the edited video and all the embeddings of editing components in the embedding queue.
% ====================================================

\section{Experiment}
\label{sec:experiment}
% We conduct experiments on two datasets for editing components, \ie, the proposed Edit3K, and AutoTransition \cite{shen2022autotransition}. 

%\subsection{Evaluation Metrics}
%\label{sec:evaluation}
%We evaluated our method on both the editing components retrieval task and the transition recommendation task. 
%For editing components retrieval, we randomly select one video rendered with a certain editing component and one pair of raw materials as the query. Then we randomly select a set of videos rendered with another pair of materials and all editing components in the evaluation set as the reference set. If the retrieved video uses the same editing component as the query video, this query is considered correct. We report the average Recall at k (denoted as R@k) for all queries. 
%The user study reports the average satisfactory rate of all retrieved results for all methods to directly evaluate the embedding distribution of all editing components. 
%For the transition recommendation task, we follow AutoTransition \cite{shen2022autotransition} to report R@k and mean rank for evaluation. 
\begin{figure*}[!htbp]
	\centering %trim=0 200 0 70, clip,
	\includegraphics[width=0.95\linewidth]{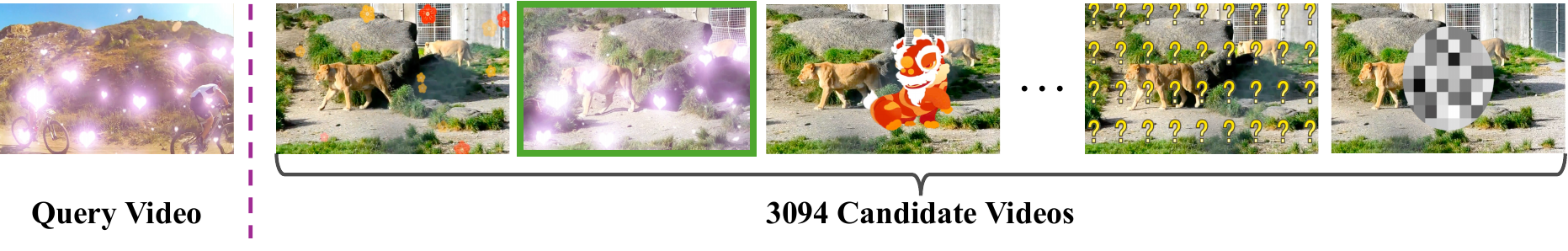}
	\caption{An example of editing components retrieval, which includes the query video and 3094 candidate videos. The green box indicates the ground truth of this query video. Best viewed on screen with zoom-in.}
	\label{fig:retrieval_task}
\end{figure*}

\begin{table*}[!htbp]
\centering
\resizebox{\linewidth}{!}{
\begin{tabular}{l | cccccccccccc| cc}
\rowcolor{mygray}
\specialrule{1.5pt}{0pt}{0pt}
 & \multicolumn{2}{c}{\textit{Video Effect}} & \multicolumn{2}{c}{\textit{Animation}} & \multicolumn{2}{c}{\textit{Transition}} & \multicolumn{2}{c}{\textit{Filter}} & \multicolumn{2}{c}{\textit{Sticker}} & \multicolumn{2}{c}{\textit{Texts}}  & \multicolumn{2}{|c}{\textit{Avg.}} \\
 \rowcolor{mygray}
 & R@1 & R@10 & R@1 & R@10 & R@1 & R@10 & R@1 & R@10 & R@1  & R@10 & R@1 & R@10 & R@1 & R@10 \\ 
 \hline
 % Classification & - & - & - & - & - & - & - & - & - & - & - & - & - & - \\
 VideoMAE \cite{tong2022videomae} & 10.4 & 16.1 & 21.6 & 29.1 & 15.0 & 23.8 & 4.9 & 6.2 & 5.2 & 5.3 & 6.8 & 6.8 & 10.7 & 14.5 \\ 
 % \redtext{VideoMoCo} \cite{pan2021videomoco} &4.1&9.2&5.0&9.9&2.6&6.9&1.3&7.9&0.7&4.8&1.1&2.7&2.5&6.9 \\
 VideoMoCo \cite{he2020momentum,pan2021videomoco} & 37.3 & 72.8 & 27.3 & 61.0 & 19.6 & 47.8 & 18.4 & 44.1 & 60.3 & 94.9 & 38.6 & \textbf{93.8} & 33.6 & 69.1 \\
Ours (Baseline) & 33.5 & 71.5 & 22.4 & 68.4 & 30.0 & 76.2 & 15.6 & 40.6 & 56.6 & 93.9 & 28.8 & 85.4 & 31.2 & 72.7 \\
Ours & \textbf{55.3} & \textbf{80.6} & \textbf{51.6} & \textbf{81.8} & \textbf{54.6} & \textbf{85.3} & \textbf{22.5} & \textbf{48.0} & \textbf{85.8} & \textbf{98.2} & \textbf{46.0} & 89.7 & \textbf{52.6} & \textbf{80.6} \\ 
\specialrule{1.5pt}{0pt}{0pt}
\end{tabular}
}
% \vspace{-0.1cm}
\caption{Comparison with state-of-the-art methods on Edit3K dataset for editing components retrieval in terms of R@k (\%).}
\label{tab:retrieval}
\end{table*}

\subsection{Implementation Details}
\label{sec:implementation}
We implement our method using PyTorch \cite{paszke2019pytorch}. We use CLIP-B/32~\cite{radford2021learning} pre-trained weights to initialize the spatial encoder and the remaining modules are all randomly initialized. We follow BERT \cite{devlin2018bert} to implement the self-attention and cross-attention transformer block. For input videos, we uniformly sample $N_v=16$ frames from each video, and the height $H$ and width $W$ of each frame is resized to $224 \times 224$. We train our model with Adam \cite{kingma2014adam} optimizer and cosine \cite{loshchilov2016sgdr} learning rate scheduling. The batch size $N_b$ per GPU is set to 8 by default and all models are trained for 20 epochs. It takes 16.7 hours to train our representation model on 8 Nvidia A100 GPUs. The learning rate of the spatial encoder is set to $1e-6$, and we use $1e-5$ for the rest of the model. The numbers of layers in the temporal encoder and embedding decoder, \ie, $N_t, N_d$, are both set to $2$. For both the encoder and the decoder, the number of attention head is set to $8$ and the hidden dimension is $512$. The total number of raw material videos and editing components, \ie, $N_m, N_e$, are $200$ and $3094$, respectively. The temperature parameter $\tau$ in Eq. \ref{eq:loss_1}, \ref{eq:loss_2} is set to $0.7$. 

\subsection{Editing Components Retrieval}
\label{sec:exp_main}

% In this section, we compare our method with state-of-the-art video representation learning methods, including self-supervised and supervised representation learning methods. For self-supervised representation learning, recent works fall into two categories, \ie, contrastive learning \cite{he2020momentum} and masked autoencoder \cite{he2022masked}. Therefore, we select two best-performing methods for comparison, \ie, VideoMoco \cite{pan2021videomoco} and VideoMAE \cite{tong2022videomae}. We re-train VideoMoco and VideoMAE on Edit3K dataset with their official code. As for supervised representation learning, early works \cite{kay2017kinetics,feichtenhofer2019slowfast} typically relied on classification tasks, and we compare our method with classification on both closed-set and open-set settings. The classification method is implemented by replacing the InfoNCE loss in Sec.~\ref{sec:method-loss} with cross-entropy loss on $3,094$ editing component classes. The output feature of the layer before classification is used as the embedding for retrieval.

In this section, we conduct experiments on the editing component retrieval task to evaluate how the embeddings of videos rendered from the same editing component and different materials are clustered. Specifically, we randomly select one video rendered with a certain editing component and one pair of raw materials as the query. Then we randomly select a set of videos rendered with another pair of materials and all editing components in the evaluation set as the candidate set. If the retrieved video uses the same editing component as the query video, this query is considered correct (the example is shown in Fig. \ref{fig:retrieval_task}). We report the average Recall at k (denoted R@k) for all queries.

We compare our method with state-of-the-art video self-supervised representation learning methods. Recent works fall into two categories, \ie, contrastive learning \cite{he2020momentum} and masked autoencoder \cite{he2022masked}. Therefore, we select two best-performing methods for comparison, \ie, VideoMoco \cite{pan2021videomoco} and VideoMAE \cite{tong2022videomae}. The original contrastive loss of VideoMoco \cite{pan2021videomoco,he2020momentum} is conducted between different frame content which does not work well on Edit3K dataset, we thus implemented a tailored version using our contrastive loss between editing components. We also re-train VideoMAE on Edit3K dataset with their official code. 

% VideoMoCo差，我们基于MoCo重新实现了VideoMoCo
%Our method achieves significantly better performance than other state-of-the-art methods.
Tab.~\ref{tab:retrieval} shows the detailed performance of six types of editing components, as well as the average accuracy. The tailored version of VideoMoCo performs on par with our baseline model, and our method performs significantly ($19\%$) better in terms of average R@1, indicating the superiority of the proposed guidance architecture. VideoMAE does not work well on Edit3K because it is optimized with reconstruction loss which does not distinguish the raw material and editing component in the video frames.

\subsection{Editing Components Representation Distribution}
\label{sec:user}
%We conduct a user study to directly evaluate the embedding distribution for all editing components, \ie, whether the clustered editing components are visually similar to each other.
Sec. \ref{sec:exp_main} has verified that the embeddings of video with the same editing components are well clustered in the embedding space, we compute their embedding center as the embedding of each editing component and evaluate the distribution across different editing components. Ideally, visually similar editing components should be close to each other in the embedding space. We split the $3,094$ editing components of Edit3K in half for open-set setting (details in supplementary material), resulting in $1,547$ editing components for training and the other half for evaluation. After training is finished, we first compute the embedding for the $1,547$ unseen editing components in the evaluation set, where the embedding of each editing component is computed as the center of all video embeddings corresponding to this specific editing component. We randomly select $27$ editing components from each of the $6$ major types as queries. Then, each query performs a search among the embedding of all the editing components except the query itself. The top-1 retrieved result is used for the user study, resulting in $162$ query-result pairs. We follow this process to generate results for $4$ methods, \ie, `Random', `VideoMoCo', `Ours (Baseline)', and `Ours'. In total, we get $648$ query-result pairs for evaluation. 
\begin{figure*}[!htbp]
    % \vspace{-0.7cm}
	\centering %trim=0 200 0 70, clip,
	\includegraphics[width=0.95\linewidth]{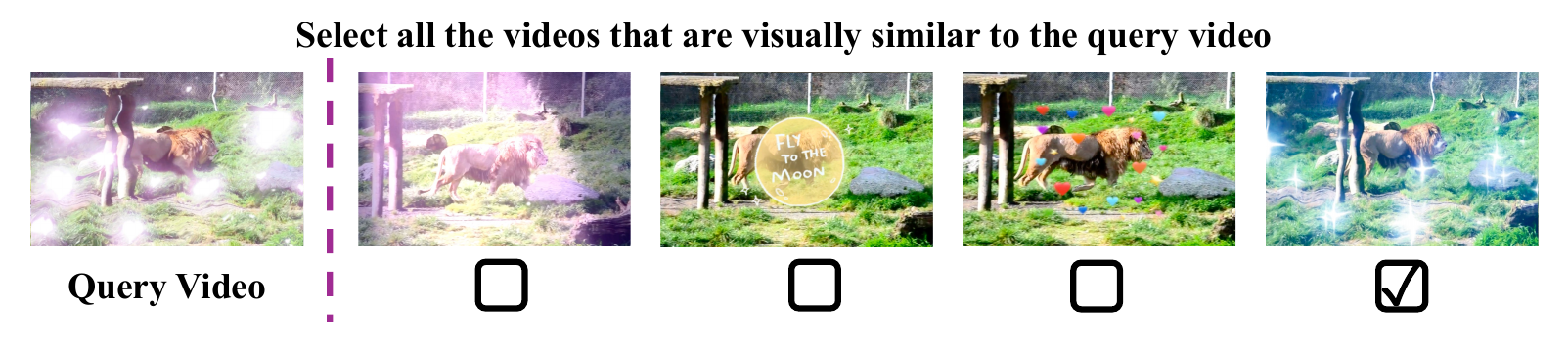}
	\caption{An example from the user study. The users are asked to select all the videos that are visually similar to the query video. Best viewed on screen with zoom-in.}
	\label{fig:interface}
\end{figure*}
\begin{table*}[!htbp]
\centering
% V.E., Anim., Tran., Filt., Stic. denote video effect, animation, transition, filter, and sticker.
% \renewcommand{\arraystretch}{1.1}
% \resizebox{0.6\textwidth}{!}{
\begin{tabular}{m{2.3cm}|m{1.3cm}m{1.3cm}m{1.3cm}m{1.3cm}m{1.3cm}m{1.3cm}|m{1.3cm}}
%{l | r r r r r r | r}
\rowcolor{mygray}
\specialrule{1.5pt}{0pt}{0pt}
  Method  & V. E. & Anim. & Tran. & Filt. & Stic. & Text & Avg. \\ 
    \hline
     Random  & $\,\,\,0.9$ & $\,\,\,0.9$ & $\,\,\,1.6$ & $\,\,\,0.9$ & $\,\,\,0.8$ & $\,\,\,1.3$ & $\,\,\,1.1$ \\
     VideoMoCo & $45.7$ & $49.1$ & $\textbf{69.5}$ & $37.5$ & $40.1$  & $52.3$ & $49.0$ \\
     Ours (Baseline) & $37.4$ & $59.4$ & $67.9$ & $39.7$  & $37.6$ & $43.2$ & $47.5$ \\
     Ours & $\textbf{61.3}$ & $\textbf{67.9}$ & $67.1$ & $\textbf{46.6}$ & $\textbf{62.4}$ & $\textbf{63.9}$ & $\textbf{61.5}$ \\
\specialrule{1.5pt}{0pt}{0pt}
\end{tabular}%}
\caption{The satisfactory rate ($\%$) of different methods in user study. V.E., Anim., Tran., Filt., Stic. denote video effect, animation, transition, filter, and sticker.}
\label{tab:user}
\end{table*}
\begin{figure*}[!htbp]
    % \vspace{-5cm}
	\centering %trim=0 200 0 70, clip,
	\includegraphics[width=0.95\linewidth]{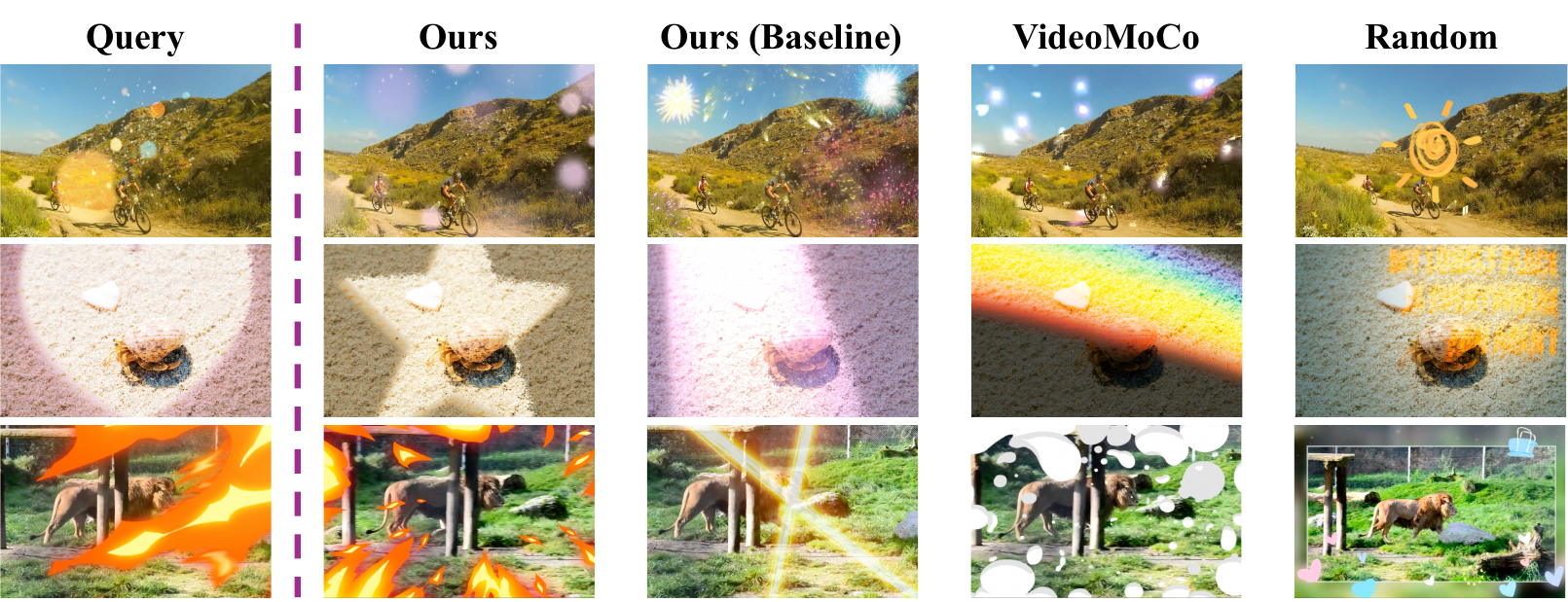}
	\caption{The top-1 similar editing components corresponding to the query editing component using different methods. The similarity is computed based on the embedding centers of editing components and we show one frame of a randomly selected video corresponding to each editing component for visualization. Best viewed with zoom-in.}
	\label{fig:user_study_res}
\end{figure*}

The users are given 5 videos in a row, where the left column is the video for the query editing component, and the other columns show the example videos for the retrieved editing components corresponding to the $4$ methods (an example is shown in Fig.~\ref{fig:interface}). The users are asked to select the videos that are visually similar to the query and each query is evaluated by at least $10$ users. The selected results are considered satisfactory and we compute the satisfactory rate of different methods on all queries. As shown in Tab.~\ref{tab:user}, `Ours' significantly outperforms the other three methods in terms of average satisfactory rate, indicating the effectiveness of the proposed guidance architecture and additional loss on embedding distribution. The result is also consistent across different editing component categories. Qualitative results is provided in Fig. \ref{fig:user_study_res}. The t-SNE \cite{van2008visualizing} visualization for the embedding space is provided in the \textbf{supplementary material}.

\begin{table*}[!htbp]
\centering
\renewcommand{\arraystretch}{1.1}
\resizebox{\linewidth}{!}{
\begin{tabular}{l | cccccccccccc | cc}
\rowcolor{mygray}
\specialrule{1.5pt}{0pt}{0pt}
\multirow{2}{*}{} & \multicolumn{2}{c}{\textit{Video Effect}} & \multicolumn{2}{c}{\textit{Animation}} & \multicolumn{2}{c}{\textit{Transition}} & \multicolumn{2}{c}{\textit{Filter}}  & \multicolumn{2}{c}{\textit{Sticker}} & \multicolumn{2}{c|}{\textit{Texts}}  & \multicolumn{2}{c}{\textit{Avg.}} \\
\rowcolor{mygray}
 & R@1 & R@10 & R@1 & R@10 & R@1 & R@10 & R@1 & R@10 & R@1  & R@10 & R@1 & R@10 & R@1 & R@10 \\ \hline
Baseline & 33.5 & 71.5 & 22.4 & 68.4 & 30.0 & 76.2  & 15.6 & 40.6 & 56.6 & 93.9 & 28.8 & 85.4 & 31.2 & 72.7 \\
Baseline + $\mathcal{L}_{q}$ & 40.3 & 76.4 & 32.7 & 69.2 & 30.9 & 72.9  & 20.8 & 39.8 & 58.8 & 94.6 & 36.1 & 90.2 & 36.6 & 73.8 \\
Baseline + $\mathcal{L}_{q}$ + GT & 43.2 & 75.9 & 37.4 & 70.9 & 39.2 & 76.4 & 20.8 & 44.3 & 58.1 & 95.1 & 41.5 & \textbf{92.4} & 40.0 & 75.8\\
Baseline + $\mathcal{L}_{q}$ + GD & 51.0 & 78.6 &  41.0 & \textbf{82.0} & 49.8 & \textbf{87.5} & 21.0 & 44.9 & 79.2 & 97.7 & \textbf{52.1} & 90.3 & 49.0& 80.2 \\
Baseline + $\mathcal{L}_{q}$ + GT + GD & \textbf{55.3} & \textbf{80.6} & \textbf{51.6} & 81.8 & \textbf{54.6} & 85.3 & \textbf{22.5} & \textbf{48.0} & \textbf{85.8} & \textbf{98.2} & 46.0 & 89.7 & \textbf{52.6} & \textbf{80.6} \\
% Baseline + $\mathcal{L}_{q}$ + FG + QG + Raw & 56.8 & 73.5 & 59.2 & 83.1 & 89.6 & 95.9 & 11.8 & 33.8 & 43.3 & 66.0 & 39.6 & 84.7 & 50.1 & 72.8 \\
\specialrule{1.5pt}{0pt}{0pt}
\end{tabular}}
\caption{Ablations study for different modules of our method on Edit3K for editing components retrieval in terms of R@k (\%). $\mathcal{L}_{q}$ is the embedding queue loss. `GT' and `GD' represent the guided tokens and guided decoder.}
\label{tab:ablations}
\end{table*}

\begin{figure*}[!htbp]
\begin{minipage}[b]{0.34\linewidth}
\centering
\begin{tabular}{ccc}
\rowcolor{mygray}
\specialrule{1.5pt}{0pt}{0pt}
Method   & R@1 & R@10 \\ \hline
None & 36.6 & 73.8 \\
K-Means & 52.1 & 79.9 \\
Default 6 Types  & \textbf{52.6} & \textbf{80.6} \\
\specialrule{1.5pt}{0pt}{0pt}
\end{tabular}
\captionof{table}{Ablation of different methods for generating embedding centers on Edit3K for editing components retrieval in terms of R@k (\%).}
\label{tab:kmean}
\end{minipage}
\hspace{0.7cm} % Optional: adjust the horizontal spacing between the figure and table
\begin{minipage}[b]{0.58\linewidth}
\centering
\includegraphics[width=.8\linewidth]{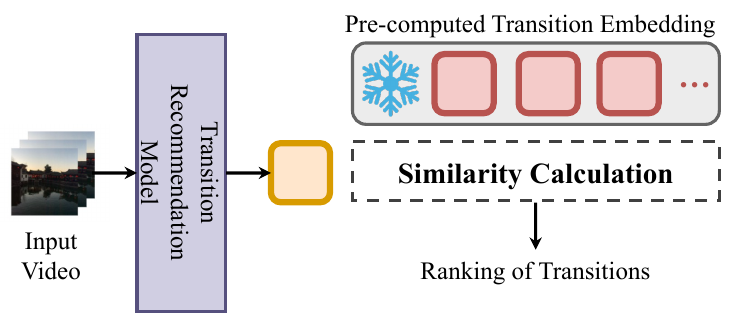}
\caption{An overview of the transition recommendation pipeline.}
\label{fig:recommendation}
\end{minipage}
\end{figure*}

\begin{table*}[!htbp]
\centering
% \renewcommand{\arraystretch}{1.1}
% `Pretrain Emb' denotes the dataset used when pretraining the method. `Train Rec' denotes the dataset used when training the transition recommendation model.
% \setlength{\tabcolsep}{10.5pt}
\begin{tabular}{c cccc c}
\rowcolor{mygray}
\specialrule{1.5pt}{0pt}{0pt}
%& Train Rec & \multirow{4}{*}{AutoTransition}
\multicolumn{2}{c}{\textbf{Transition Embedding}} &  & \multicolumn{3}{c}{\textbf{Performance}} \\
\cline{1-2} \cline{4-6}
\rowcolor{mygray}
Method & Dataset  & & R@1 & R@5 & Rank ($\downarrow$) \\ \hline
Random \cite{shen2022autotransition} & -  & & 25.67 & 66.30 & 5.65 \\
Classification\cite{shen2022autotransition} & AutoTransition & & 28.06 & 66.85 & 5.48  \\
Ours-A & AutoTransition & & 28.89 & 67.12 & 4.86 \\ 
Ours-E & Edit3K  & & \textbf{29.24} & \textbf{67.32} & \textbf{4.56} \\ 
\specialrule{1.5pt}{0pt}{0pt}
\end{tabular}
\caption{Comparison of different transition embeddings on the AutoTransition \cite{shen2022autotransition} test set in terms of R@k (\%) and mean rank. \textit{All the rows use the same recommendation model training pipeline from \cite{shen2022autotransition} and the only difference is the pre-computed fixed embedding for the transitions.}}
\label{tab:transition}
\end{table*}

\subsection{Ablation Study}
\label{sec:ablation}
\noindent \textbf{Different Modules.} 
We conduct ablation study in Tab.~\ref{tab:ablations} by removing different modules to demonstrate the effectiveness of each module. The `Baseline' model with only spatial-temporal encoder and in-batch InfoNCE loss achieves an average R@1 score of $31.2\%$. Adding the embedding queue loss term (`Baseline + $\mathcal{L}_{q}$') brings $5.4\%$ improvement on the R@1 accuracy, indicating the effectiveness of the $\mathcal{L}_{q}$ in Eq. \ref{eq:loss_2}. The R@1 is further improved by $3.4\%$ when adding guidance tokens in the spatial encoder (`Baseline + $\mathcal{L}_{q}$ + GT'). Instead, deploying the guided embedding decoder (`Baseline + $\mathcal{L}_{q}$ + GD') also gains $12.4\%$ improvement on R@1 over `Baseline + $\mathcal{L}_{q}$'. The performance is further improved when all the modules are employed (`Baseline + $\mathcal{L}_{q}$ + GT + GD'), demonstrating the effectiveness of the guidance architecture.

\noindent \textbf{Embedding Centers.} Tab.~\ref{tab:kmean} shows the ablation study using different embedding centers as guidance tokens. Both `Default 6 Types' and `K-Means' perform well with a significant performance boost over `None' (no embedding centers), indicating our guidance architecture is generic for different clustering methods. The default version with 6 editing component types performs slightly better on Edit3K, but the k-means clustering centers are more flexible for other tasks.
% \redtext{When generating guidance tokens, we use the cluster centers instead of directly using thousands of embedding vectors, which not only reduces computational load, but also minimizes noise. To explore the impact of different methods for generating embedding centers on model performance, we conducted experiments using embedding centers from both the default six types and K-means clustering, as shown in Tab.~\ref{tab:kmean}.  From the table, it can be observed that the performance of the two methods is comparable, and K-means also performs well, demonstrating the generality of our architecture, which is not limited to just six types.}

\subsection{Transition Recommendation}
\label{sec:exp_auto}
We conduct experiments on transition recommendation as a downstream task to validate the generalization ability of our learned universal representation. In \cite{shen2022autotransition}, the embeddings of transitions are pre-computed with a classification model trained on AutoTransition dataset, and the embeddings are fixed during the training of the recommendation model. As shown in Fig. \ref{fig:recommendation}, we follow \cite{shen2022autotransition} to train transition recommendation models which take a raw video as input and generate ranking scores for a set of transitions. We replace the pre-computed transition embeddings with other alternative methods to validate the benefit of a strong representation on downstream tasks. We follow \cite{shen2022autotransition} to report R@k and mean rank for evaluation.

% \redtext{In the transition recommendation task, the quality of the transition embedding is crucial for the accuracy of the recommendations. To further validate the quality of the embeddings we generated, we conducted verification on the transition recommendation task. In our experiments, we solely explore different methods to generate transition embeddings, keeping the transition recommendation model and recommendation dataset (AutoTransition) consistent. We follow AutoTransition \cite{shen2022autotransition} to report R@k and mean rank for evaluation. }
\begin{figure*}[!htbp]
 \centering
 \includegraphics[trim={330 0 340 100},clip, width=\linewidth]{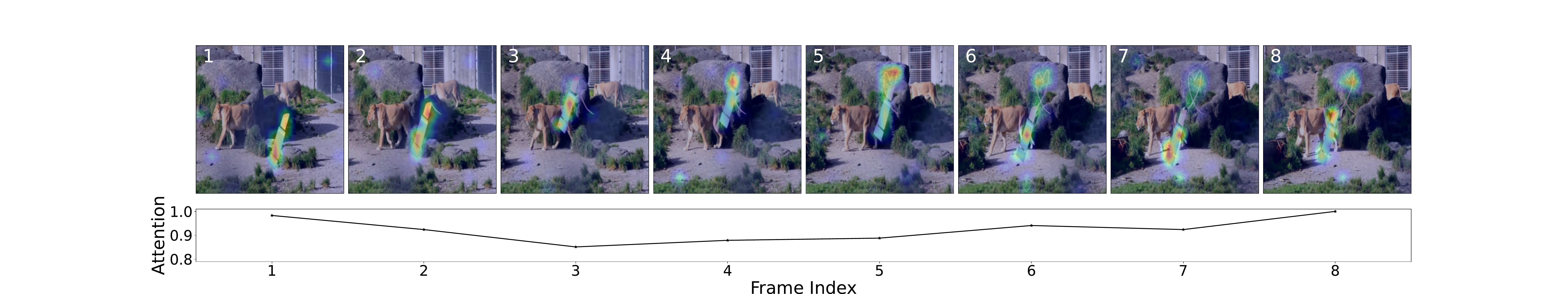} %530_531_564884.mp4_atten_8.pdf, {190 20 170 40}
 %530_531_564884.mp4_atten.pdf {410 20 340 40}
 \caption{Visualization of the attention map in the spatial and temporal encoder. The first row shows the spatial attention maps overlaid on the 8 input frames, the bottom axis shows the temporal attention value for each frame. The model focuses on the lipstick-shaped sticker for all the frames without explicit spatial supervision. Best viewed on the screen with zoom-in.}
 \label{fig:visualization}
\end{figure*}

Tab.~\ref{tab:transition} shows that the embedding of the proposed method trained on AutoTransition datasets (`Ours-A') outperforms the state-of-the-art result in \cite{shen2022autotransition} (`Classification'), indicating the superiority of the proposed architecture. Since the transitions in AutoTransition dataset are included in our Edit3K dataset, we further adopt the pre-computed embedding on Edit3K (`Ours-E') and it brings additional performance boosts, indicating the superiority of our diverse and balanced dataset. \textit{Note that all the rows use the same recommendation method and the only difference is the pre-computed transition embeddings, a slight performance boost could indicate significant improvement on representation learning}.
% The experimental results are shown in Tab.~\ref{tab:transition}. The results indicate that the universal representation learned on Edit3K can generalize well on unseen data and the proposed Edit3K dataset could serve as a good representation pre-training set for all editing components downstream tasks, \eg, recommendation, detection, etc.

% Since the transitions in AutoTransition \cite{shen2022autotransition} dataset are included in our Edit3K dataset, the embedding computed on Edit3K can be directly adopted to AutoTransition dataset. The transition recommendation pipeline in AutoTransiton \cite{shen2022autotransition} takes the pre-trained embedding of all transitions as part of the inputs and the model learns to recommend transitions based on input video frames and audio. We use the same training and evaluation pipeline as AutoTransition \cite{shen2022autotransition}, and we compare the performance of different representation learning methods by simply changing the pre-trained embedding.

\subsection{Attention Visualization}
Fig.~\ref{fig:visualization} shows both the spatial and temporal attention map of our model by visualizing the multi-head self-attention/cross-attention module of the last transformer layer in the spatial encoder and temporal encoder, respectively. The output of our model highly depends on the highlighted regions. In Fig. \ref{fig:visualization}, a lipstick-shape sticker is added in the middle of the scene with moving animals. Although our model is not trained with explicit segmentation supervision, our model learns to focus on the sticker and ignore the background changes when both the sticker and the background scene are moving. The temporal attention values of the eight frames are similar to each other, as the sticker appears on all the frames. More results can be seen in the \textbf{supplementary material}.

% ====================================================

\section{Conclusion}
\label{sec:conclusion}
We introduce the first large-scale editing components dataset with $618,800$ rendered videos covering 6 major types of editing components, which can benefit related downstream tasks. We further propose a novel embedding guidance architecture and a specifically designed contrastive loss for editing component representation learning. It achieves state-of-the-art performance on both editing components retrieval and a major downstream task, \ie transition recommendation. The user study also demonstrates that the learned embedding distribution of our model is better than that of existing methods. 

One limitation is that our current model uses low frames per second (FPS), which cannot handle extremely fast motion or appearance change. In addition, some editing components may have very small changes on the raw videos which is very challenging for our model to recognize without raw video as input. How to efficiently take advantage of the raw videos is worth exploring in the future. 

With the rapid growth of video-sharing platforms, edited videos have become one of the major data sources on the Internet. This work serves as a solid step toward understanding video editing for better recommendation, security, etc. The authors do not foresee any negative social impact.

%Limitation and societal impact are included in supplementary material. 
% \Sijie{Maybe add limitation and societal impact here or in supplementary material.}

{
    \small
    \bibliographystyle{ieeenat_fullname}
    \bibliography{main}
}

\end{document}